\documentclass[runningheads]{llncs}
\usepackage{xurl}
\usepackage{graphicx}
\usepackage{subcaption}
\usepackage{tikz}
\usepackage{float}
\usetikzlibrary{calc}
\usepackage{nicematrix}
\usepackage[utf8]{inputenc}
\usepackage[T5]{fontenc}

\usepackage{booktabs}
\usepackage{multirow}
\usepackage{color, colortbl}

\usepackage{pbox}

\usepackage{mathrsfs}
\usepackage[lined, ruled, linesnumbered, noend, scleft, nofillcomment]{algorithm2e}
\SetKwRepeat{Do}{do}{while}
\usepackage{amsmath,amsfonts}
\usepackage{algorithmic}
\usepackage{graphicx}
\usepackage{textcomp}

\usepackage{amsmath}

\usepackage{newtxmath}
\usepackage{parcolumns}
\begin{document}
\title{REBot: From RAG to CatRAG with Semantic Enrichment and Graph Routing}
\titlerunning{REBot: Category-guided GraphRAG for Academic Regulation Advising}
%
%
\author{
Thanh Ma\and
Tri-Tam La\and
Lam-Thu Le Huu\and
Minh-Nghi Nguyen\and
\\Khanh-Van Pham Luu
}
\authorrunning{Thanh Ma et al.}
%
\institute{
\email{mtthanh@ctu.edu.vn\\
\{tamb2203579,thub2206018,nghib2203570,vanb2203592\}@student.ctu.edu.vn}
}
\maketitle
\begin{abstract}


Academic regulation advising is vital for helping students interpret and comply with institutional policies, yet building effective systems requires domain-specific regulatory resources. To address this challenge, we propose REBot, an LLM-enhanced advisory chatbot powered by CatRAG, a hybrid retrieval–reasoning framework that integrates RAG with GraphRAG. We introduce CatRAG that unifies dense retrieval and graph-based reasoning, supported by a hierarchical, category-labeled knowledge graph enriched with semantic features for domain alignment. A lightweight intent classifier routes queries to the appropriate retrieval modules, ensuring both factual accuracy and contextual depth. We construct a regulation-specific dataset and assess REBot on classification and question-answering tasks, achieving state-of-the-art performance with an F1-score of 98.89\%. Finally, we implement a web application that demonstrates the practical value of REBot in real-world academic advising scenarios.

\end{abstract}
%
%
\section{Introduction}
Addressing student inquiries regarding university regulations remains a persistent challenge for many Vietnamese universities and institutes. At Can Tho University (CTU), such information is primarily disseminated through official sources including the student handbook, departmental websites, and PDF documents. Consequently, students often face difficulties in locating precise answers, leading to uncertainty about critical academic procedures and policies. Common questions include: 'What is the maximum study duration?', 'What are the requirements for receiving a university scholarship?', or 'How can I change my major?' To find answers, students may either consult multiple documents or seek assistance from the relevant department, both of which can be time-consuming and inefficient. Frequent policy updates further complicate maintaining awareness of the latest regulations.

Artificial intelligence (AI) has been widely adopted in chatbot-based advisory systems. In Vietnam, Phuoc et al.~\cite{hueuniversity} employed Rasa to build a regulation management chatbot capable of intent detection, entity extraction, and dialogue management; while effective for FAQs, its heavy reliance on a small dataset limited robustness for unseen queries. Similarly, Luong and Luong~\cite{luong2025chatbot} developed an NLP/ML chatbot for student services, providing fast responses but struggling with policy-specific reasoning. Recent advances in large language models (LLMs)~\cite{YAO2024100211,NAZIR2023100022,inproceedings} such as ChatGPT and Gemini overcome these issues by modeling complex semantics and sustaining multi-turn dialogue, offering valuable insights for academic advisory systems. Complementary techniques have also emerged: Retrieval-Augmented Generation (RAG) improves factual grounding but falters with ambiguous queries, whereas Knowledge Graphs (KGs) capture structured policy relations but lack generative flexibility. To address these gaps, we integrate both in GraphRAG, achieving precise, context-aware, and explainable responses to complex regulation-related questions.

We present REBot, an LLM-powered chatbot that offers CTU students \cite{ma2024racos} fast, accurate, and up-to-date answers on academic regulations. At its core is CatRAG, a hybrid framework combining RAG for factual reliability with GraphRAG for structured contextual reasoning. To meet the dual need for direct responses and explanatory context, CatRAG integrates NER for semantic enrichment and a routing classifier for precise subgraph selection, enhancing both accuracy and efficiency. Our contributions include: (1) a curated CTU Academic Regulation Dataset optimized for retrieval; (2) a three-tier Knowledge Graph linking categories, chunks, and entities for explainable reasoning; (3) an intent classifier for query routing; (4) CatRAG, a category-guided extension of RAG with new Construct and Query algorithms; and (5) the REBot web application, enabling real-time access to regulation guidance.

The primary goal of GraphRAG is not simply retrieving top-ranked answers, as in conventional RAG, but supplying surrounding contextual knowledge that enriches responses and yields more comprehensive advice. Yet, applied in isolation, GraphRAG often underperforms standard RAG \cite{han2025rag}. Our approach therefore seeks both to generate accurate answers and to integrate relevant context, enhancing the overall quality of academic advising.

\section{Background}

\subsection{Academic Regulations Inquiry}

Academic regulations vary across universities, each with distinct administrative structures and formats. At Can Tho University, regulations are issued under official decisions and published online\footnote{\url{https://dsa.ctu.edu.vn/noi-quy-quy-che/quy-che-hoc-vu.html}}, yet the documents remain fragmented, unstructured, and written in formal administrative language, making efficient retrieval difficult for students. To address this, we propose an AI-powered chatbot tailored for CTU, leveraging four core techniques: query-oriented text classification, Named Entity Recognition (NER), Retrieval-Augmented Generation (RAG), and GraphRAG. These components are elaborated in the following sections.

\subsection{Query-Oriented Text Classification for Academic Regulations}

Classifying student queries into thematic categories is essential for improving access to academic regulations. While traditional methods \textit{(e.g., Naïve Bayes, KNN, SVM)}  are widely used~\cite{mironczuk2018recentrec,li2022survey}, they struggle with the linguistic complexity of natural language. In this work, we take advantage of deep learning models \textit{(e.g., fasttext)} for the classification model. Formally, given queries $\mathcal{X}$ and categories $\mathcal{Y} = \{y_1, \dots, y_k\}$, a classifier $\Phi: \mathcal{X} \to \mathcal{Y}$ maps each query to its most relevant category, enabling intent inference and accurate responses. To ensure coverage, regulatory data are extracted from dispersed PDFs and continuously updated from CTU’s official sources.

\subsection{Semantic Enrichment with NER}

Named Entity Recognition (NER) is a central task in information extraction, identifying entities such as people, organizations, and locations~\cite{goyal2018recent,8999622}. A typical NER pipeline involves segmentation, tokenization, POS tagging, and entity detection, transforming unstructured text into structured representations~\cite{huyen2016vlsp}. In this study, we highlight POS tagging as it provides syntactic cues that constrain entity search, reduce false positives, and improve accuracy~\cite{chiche2022part,minh2018featurerich}. For Vietnamese, we employ \textit{Underthesea}\footnote{\url{https://github.com/undertheseanlp/underthesea}}, a robust deep learning toolkit with efficient POS tagging~\cite{nguyen2017word}. Formally, given a token sequence \(\mathit{T} = (t_1, \dots, t_n)\), the NER function maps
$
\mathbf{e} = \mathcal{N}(\mathit{T}) = (e_1, \dots, e_n), \quad e_i \in \mathcal{E},
$ where \(\mathcal{E}\) is the entity label set. This formulation underscores POS tagging as a vital step that enhances NER efficiency and accuracy, thereby improving chatbot responses. The integration of RAG and GraphRAG as the framework’s core will be detailed in the following sections.

\subsection{Baseline/General Responses using RAG}

Large language models (LLMs), while fluent and generalizable, are prone to \textit{AI hallucination}\footnote{\url{https://cloud.google.com/discover/what-are-ai-hallucinations}}. Retrieval-augmented methods mitigate this by grounding generation in external knowledge such as corpora, PDFs, or databases. Formally, given a corpus $\mathcal{D} = \{D_1, \dots, D_n\}$ segmented into chunks $\mathcal{C} = \{c_1, \dots, c_m\}$, an embedding function $f_{\mathrm{emb}}$ maps each chunk into $\mathbb{R}^d$, producing a vector store
\(
\mathcal{V} = \{(c_j, f_{\mathrm{emb}}(c_j)) \mid c_j \in \mathcal{C}\},
\)
which supports efficient similarity-based retrieval. We adopt RAG as the primary mechanism, as it anchors responses in semantically relevant evidence, reducing hallucination and improving factual reliability~\cite{lewis2021retrievalaugmentedgenerationknowledgeintensivenlp}. This retrieval–generation paradigm provides the foundation for GraphRAG, which further extends capability with structured graph knowledge.

\subsection{Enhanced/Extended Responses using GraphRAG}

\begin{figure}[ht]
\centering
\resizebox{0.8\textwidth}{!}{
\begin{subfigure}{0.4\textwidth}
\begin{tikzpicture}[remember picture]
  \node[anchor=south west, inner sep=0] (A) at (0,0)
    {\includegraphics[width=\linewidth]{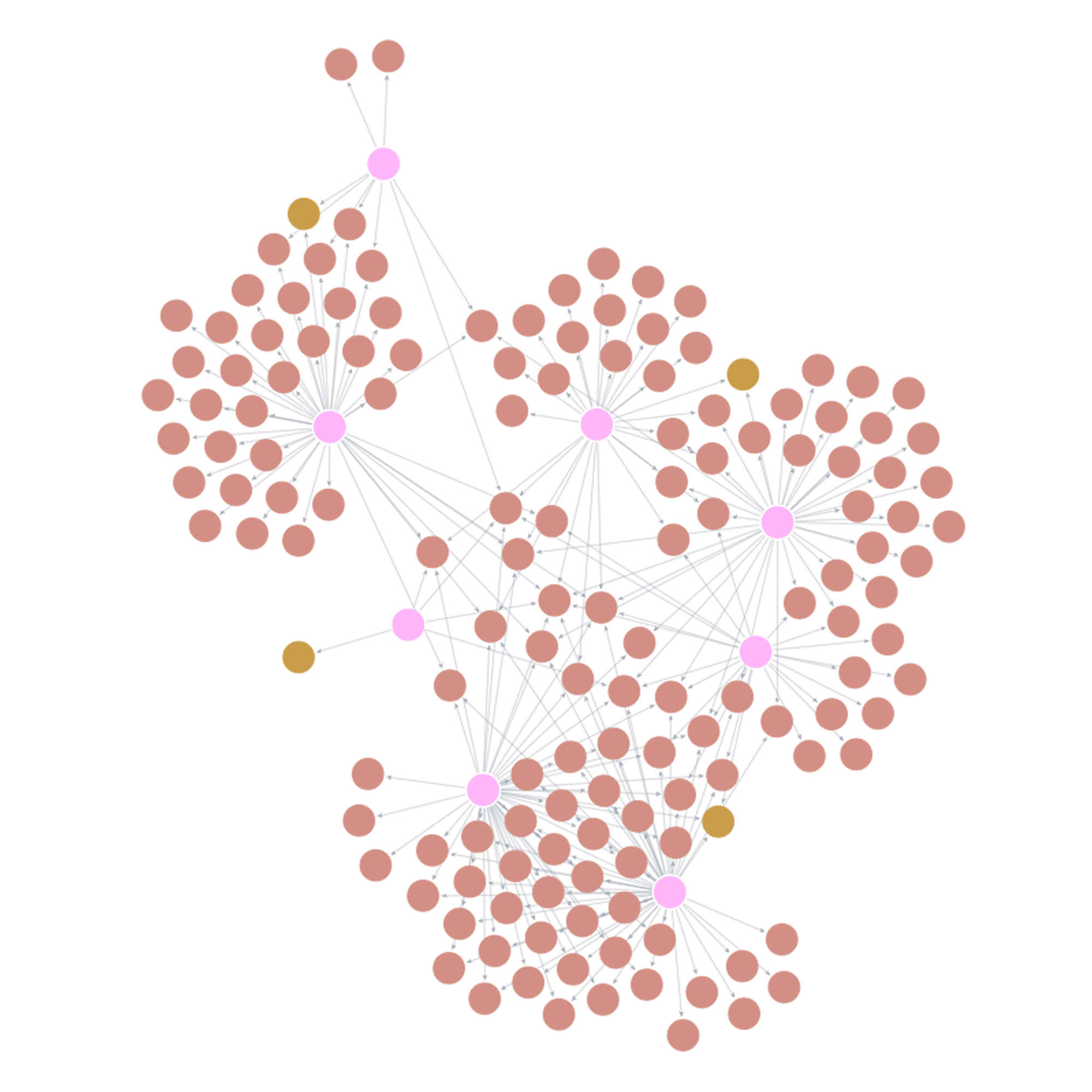}};

  \begin{scope}[x={(A.south east)}, y={(A.north west)}]
    \coordinate (ZSW) at (0.40,0.10); 
    \coordinate (ZNE) at (0.60,0.24); 
    \coordinate (ZSE) at (0.60,0.10); 
    \coordinate (ZNW) at (0.40,0.24); 

    \draw[red,thick] (ZSW) rectangle (ZNE);
  \end{scope}
\end{tikzpicture}
\label{fig:kg}
\end{subfigure}
\hfill
\begin{subfigure}{0.38\textwidth}
\begin{tikzpicture}[remember picture]
  \node[anchor=south west, inner sep=0] (B) at (0,0)
    {\includegraphics[width=\linewidth]{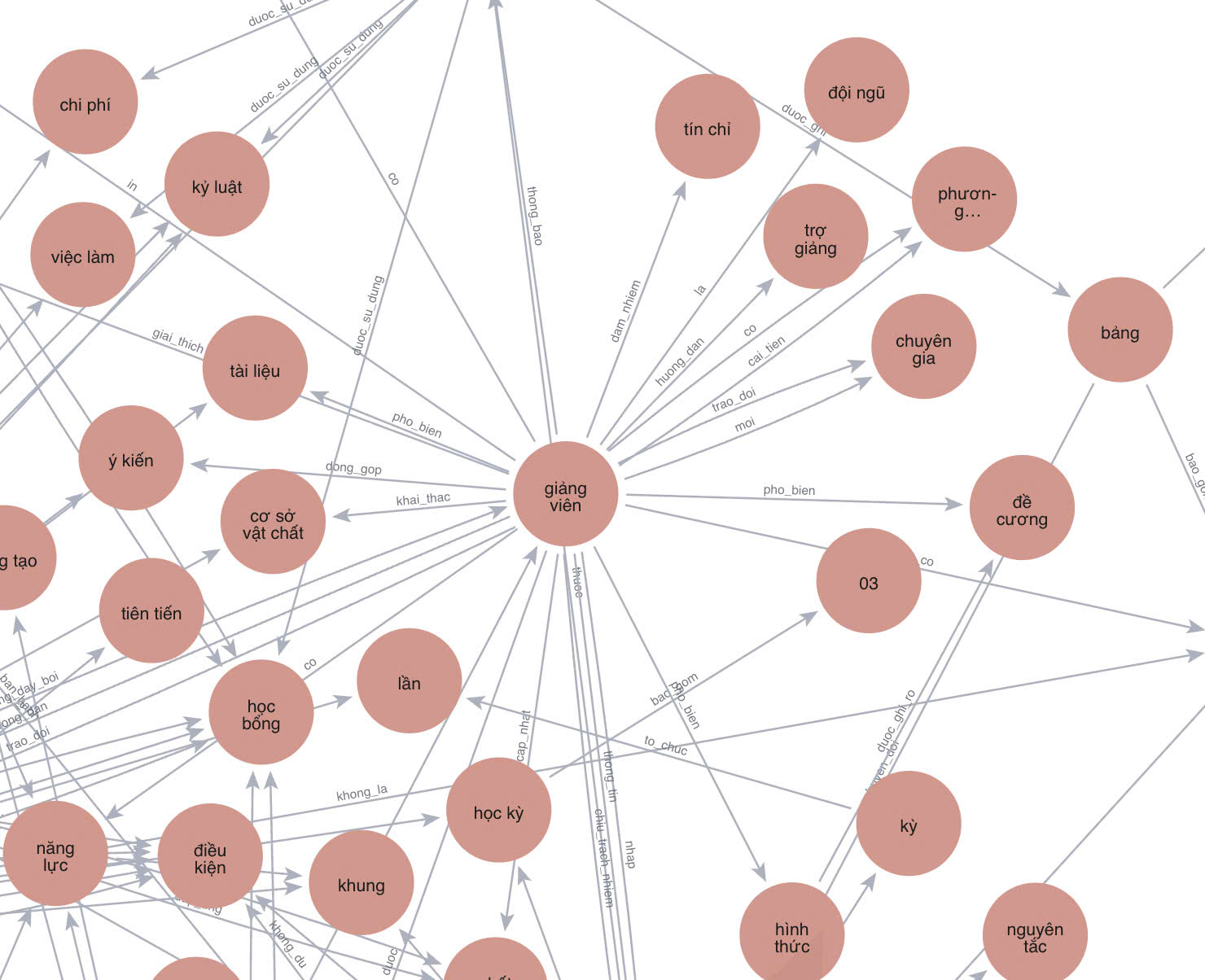}};
  \coordinate (Bwestmid) at ($(B.west)!0.5!(B.north west)$);
\end{tikzpicture}
\label{fig:nodes}
\end{subfigure}
\begin{tikzpicture}[remember picture, overlay]
  \draw[red,thick] (ZNE) -- (Bwestmid);
  \draw[red,thick] (ZSE) -- (Bwestmid);
\end{tikzpicture}
}

\caption{Knowledge graph of REBot with 250 nodes}
\label{fig:KnowledgeGraphZoom}
\end{figure}
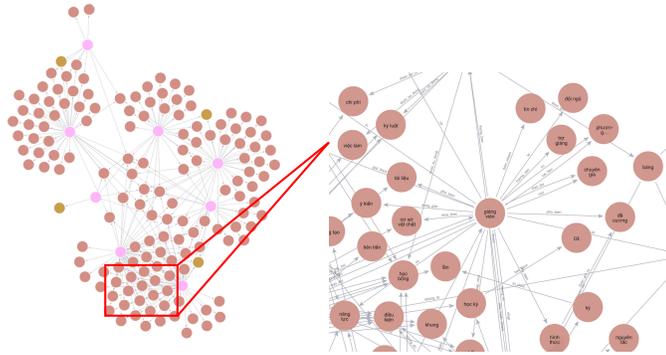

GraphRAG~\cite{han2024retrieval} extends RAG by integrating structured knowledge from knowledge graphs (KGs) into retrieval and generation. Unlike standard RAG, which relies on unstructured corpus, GraphRAG enables entity-centric reasoning and multi-hop traversal over relational structures~\cite{hu2024grag,lewis2021retrievalaugmentedgenerationknowledgeintensivenlp}. As an overview, Figure~\ref{fig:KnowledgeGraphZoom} illustrates the structure of our knowledge graph; its formal definition and role within CatRAG will be detailed in the algorithm section. This structure supports fact retrieval, symbolic reasoning, and semantic search~\cite{Hogan_2021,peng2023knowledge,pan2024unifying}, thereby enhancing factual consistency and verifiability~\cite{sun2023think,fatemi2023talk}. Hence, we adopt GraphRAG for extended responses for it enriches search with structured, entity-level knowledge, allowing deeper reasoning beyond unstructured retrieval alone.

In REBot, this design allows the system to provide domain-specific, context-aware, and verifiable responses. The subgraph retrieval ensures that answers are grounded not only in relevant documents but also in the correct entity relationships, which is essential for regulation-focused dialogue. The next section explains details about our REBot framework.

\section{REBot Framework}
We present \textbf{REBot}, a framework for precise, context-aware responses, initially built to support CTU students with academic regulations. By combining semantic query analysis with domain-specific context, it offers accurate and adaptable answers, paving the way for universal next-generation chatbot systems.

\begin{figure}[ht]
 \begin{center}
      \includegraphics[scale=0.13]{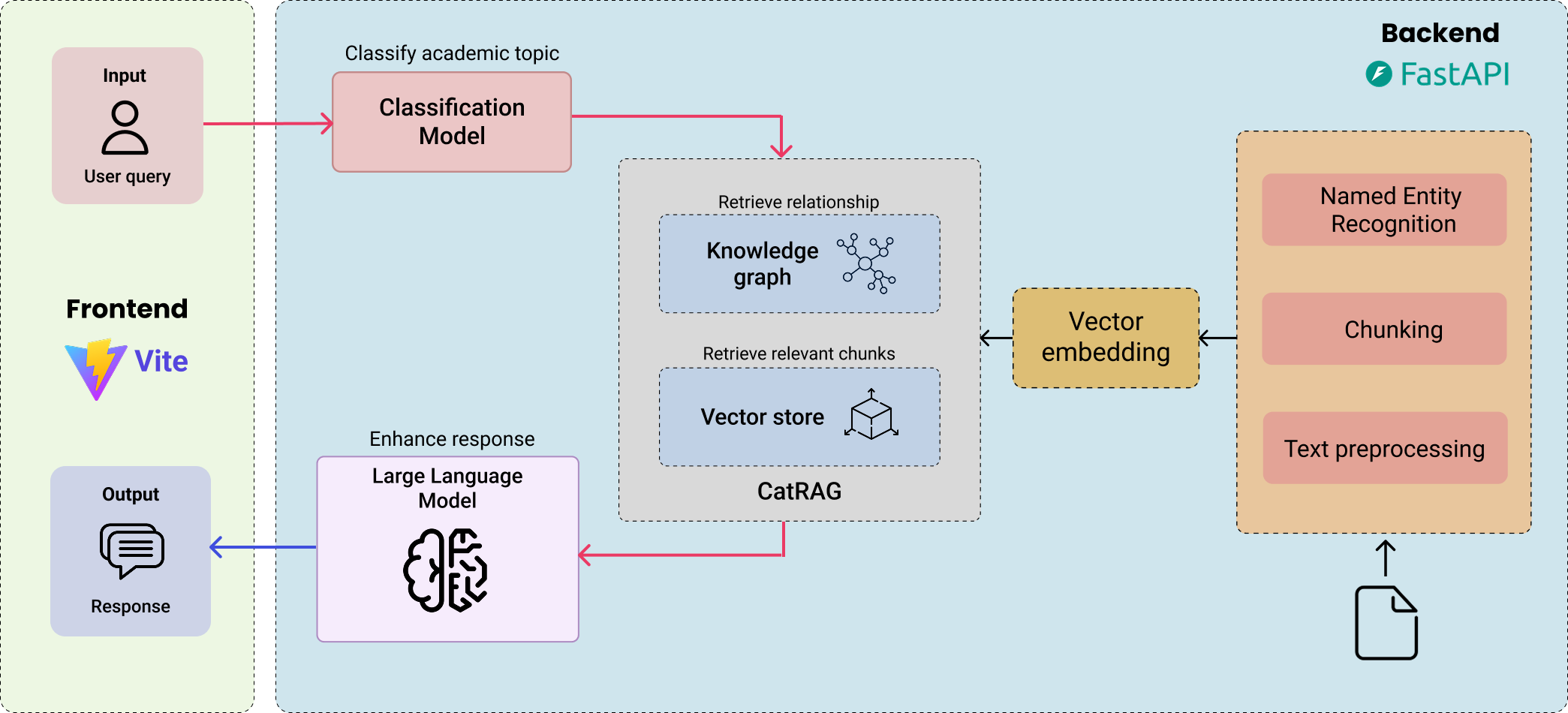}
 \end{center}
\caption{The REBot framework}
\label{ViTIPframework}
\end{figure}

The framework comprises five interrelated components: (1) \textit{Knowledge Extraction}, where PDFs are processed with Docling\footnote{\url{https://github.com/docling-project/docling}} for accurate text parsing, cleaned, segmented, and enriched via NER and relation extraction with Underthesea\footnote{\url{https://github.com/undertheseanlp/underthesea}}, then embedded using PhoBERTv2 \cite{nguyen2020phobert} for reliable Vietnamese representations; (2) \textit{Student Affairs Classification} ($\Phi$), in which queries are categorized into five regulation domains using fastText \cite{Yao2020TextCM}, chosen for its efficiency and accuracy on short Vietnamese texts; (3) \textit{REBot Knowledge Graph}, a Neo4j vector-enabled database organizing knowledge into five domain graphs with three layers (categories, contexts, entities), capturing complex student–affairs relations; (4) \textit{Vectorization and Embeddings}, where LlamaIndex\footnote{\url{https://www.llamaindex.ai/}} builds a vector index for RAG, combined with PhoBERTv2 embeddings for semantic similarity, and GraphRAG integration to unify vector retrieval with graph structures; and (5) \textit{Response Refinement}, where Mistral\footnote{\url{https://console.mistral.ai/api-keys}} and ChatGPT\footnote{\url{https://platform.openai.com/docs/api-reference}} refine retrieved content into fluent, context-aware answers.

\section{CatRAG Algorithm}
\label{compute-score}
CatRAG is the core idea of this study. It extends RAG by combining dense semantic retrieval from a vector store with symbolic reasoning over a regulation-specific knowledge graph. The prefix ``Cat'' highlights its category-guided design, where a classifier routes queries to relevant domains, i.e., enrollment, graduation, assessment (see Table \ref{lb-Dataset}). In CatRAG, documents are chunked, linked to entities, and organized into subgraphs, unifying unstructured and structured retrieval for academic advising. First of all, we formalize CatRAG by introducing the Regulation-Enriched Graph (REG), a structured representation of academic regulations. REG maps regulation chunks into entities and triplets, while the classification model $\Phi$ partitions the graph into category-specific subgraphs. Its definition is as follows:

\begin{definition}[REG]
Let $\mathcal{C} = \{c_1, \dots, c_M\}$ be a set of document chunks. A REG is a labeled graph $G = (\mathcal{E}, \mathcal{R})$, where $\mathcal{E}$ is the set of entities extracted from $\mathcal{C}$, and $\mathcal{R} \subseteq \mathcal{E} \times \mathcal{P} \times \mathcal{E}$ is the set of relations. Each relation $r \in \mathcal{R}$ is a triplet $(e_i, p, e_j)$ with predicate $p \in \mathcal{P}$ linking entities $e_i, e_j \in \mathcal{E}$. Furthermore, $G$ is partitioned into subgraphs $\{g_y\}$, each corresponding to a category label predicted by $\Phi$.
\end{definition}

Before presenting the two CatRAG algorithms, we outline the preprocessing pipeline that ensures domain-specific accuracy: (i) extracting text from PDFs with Docling and OCR, (ii) normalizing text via lowercasing, stopword removal, and LLM-based correction, and (iii) enriching semantics with a custom regulation dictionary containing 46 abbreviation–full term pairs in the academic domain. Togerther, these steps enhance entity extraction for the knowledge graph and similarity matching in the vector store. The next subsections detail \textit{CatRAG-Construct} and \textit{CatRAG-Query}.

\subsection{CatRAG-Construct Algorithm}

\textit{CatRAG-Construct} builds a hybrid knowledge base by indexing chunks in a vector store and linking them to a category-guided knowledge graph. Documents are chunked, embedded with $f_{\mathrm{emb}}$ into $\mathcal{V}$ for fast semantic retrieval, while each chunk is classified by $\Phi$ and attached to its category node in $G$ to ensure domain locality. Entities extracted via $\mathcal{N}$ are linked to chunks, and relations are added by an LLM to form typed edges, yielding explainable graph-structured evidence aligned with the vector index. This design leverages dense retrieval for recall and graph structure for precision, provenance, and reasoning, with the category layer narrowing the search space and preventing domain leakage.

\BlankLine
\begin{algorithm}[ht]
\tiny
\caption{CatRAG-Construct}
\label{alg:graphrag_unified}
\DontPrintSemicolon
\SetKwInOut{Input}{Input}
\SetKwInOut{Output}{Output}
\Input{$\mathcal{D} = \{D_1, \dots, D_n\}$: Corpus of documents; $\mathcal{C} = \{c_1,\dots,c_M\}$: Corpus of document chunks; $f_{\mathrm{emb}}: \mathbb{T} \rightarrow \mathbb{R}^d$: Embedding function mapping text to $d$-dimensional vectors; $G$: Knowledge graph; $\mathcal{V}$: Vector store; $\mathcal{N}$: Named entity recognition-POS tagging model; $\Phi$: Text classifier model;
$\mathsf{LLM}$: Large language model for relation extraction}
\Output{
    $G$, $\mathcal{V}$
}
\Begin{
    $\mathcal{V} \gets \emptyset$, $G \gets \emptyset$\;
    \For{$D \in \mathcal{D}$}{
        segment $D$ into chunks $\{c_1, \dots, c_k\}$ and add to $\mathcal{C}$\;
    }
    \For{$c \in \mathcal{C}$}{
        $\mathbf{v}_c \gets f_{\mathrm{emb}}(c_{\text{text}})$\; 
        store $(c_{\mathrm{id}}, c_{\text{text}}, \mathbf{v}_c)$ in $\mathcal{V}$\;
        
        $label_c \gets \Phi(c_{\text{text}})$\;
        attach $c$ and $v_c$ to category $label_c$ in $G$\;
        
        $T \gets \mathrm{tokenize}(c_{\text{text}})$\;
        $\mathcal{E}_c \gets \mathcal{N}(T)$\;
        \For{$e \in \mathcal{E}_c$}{
            add node $e$ to $G$; create edge $(c)\text{-[:MENTIONS]}-(e)$\;
        }
        
        $\mathcal{R}_c \gets \mathsf{LLM}.\mathrm{extract\_relations}(c_{\text{text}}, \mathcal{E}_c)$\;
        \For{$(e_i, p, e_j) \in \mathcal{R}_c$}{
            add edge $(e_i)\text{-[:RELATED\_TO \{predicate\}=p, chunk\_id}=c_{\mathrm{id}}]-(e_j)$ to $G$\;
        }
    }
    \Return{$G$, $\mathcal{V}$}\;
}
\end{algorithm}

In terms of complexity, preprocessing and chunking are linear in corpus size; embedding $M$ chunks costs $O(Md)$; classification adds $O(M\,C_{\Phi})$; NER adds $O\!\left(\sum_{c}|c|\right)$; and LLM-based relation extraction contributes $\sum_{c} C_{\mathrm{RE}}(|c|)$, typically dominant. Space usage is $O(Md)$ for the vector index and $O(|V|+|E|)$ for the graph. Practical considerations include choosing chunk size to balance recall vs.\ redundancy, rate-limiting LLM relation extraction, and enforcing schema/typing to control graph sparsity and noise.

\subsection{CatRAG-Query Algorithm}

The core idea of \textit{CatRAG-Query} (Algorithm~\ref{alg:graphrag_query}) is to unify vector-based retrieval with category-guided graph retrieval, balancing recall efficiency with domain-specific precision. A user query is embedded and classified into an academic category, then matched against top-$k_{\mathrm{vec}}$ results from the global vector store and top-$k_{\mathrm{graph}}$ chunks from the corresponding knowledge subgraph. Entities and relations are expanded to enrich evidence, which is passed with the query to an LLM for final answer generation. This hybrid design addresses the limits of vector retrieval (low explainability) and pure graph retrieval (over-restrictiveness), yielding context-aware, accurate, and explainable responses.

\BlankLine
\begin{algorithm}[ht]
\tiny
\caption{CatRAG-Query}
\label{alg:graphrag_query}
\DontPrintSemicolon
\SetKwInOut{Input}{Input}
\SetKwInOut{Output}{Output}
\Input{ $q$: user query,  $f_{\mathrm{emb}}: \mathbb{T} \rightarrow \mathbb{R}^d$: embedding function mapping text to $d$-dimensional vectors, $\mathcal{V}$: vector store, $G$: knowledge graph, partitioned into labeled subgraphs $\{g_y\}$, $\Phi$: classification model mapping text to labels,  $\mathsf{LLM}$: large language model for context synthesis, $k_{\mathrm{vec}}$: number of top vector matches.}
\Output{
    $rep$: generated answer
}
\Begin{
    $\mathbf{v}_q \longleftarrow f_{\mathrm{emb}}(q)$\;
    $label_q \longleftarrow \Phi(q)$\;

    $\mathcal{C}_{\mathrm{vec}} \longleftarrow \mathrm{TopK}(\mathcal{V}, \mathbf{v}_q, k_{\mathrm{vec}})$\;

    $\mathcal{C}_{\mathrm{graph}} \longleftarrow \emptyset$,
    $\mathcal{E}_{\mathrm{graph}} \longleftarrow \emptyset$,
    $\mathcal{R}_{\mathrm{graph}} \longleftarrow \emptyset$\;

    \For{$g_y \in G$}{
        \If{$y = label_q$}{
            $\mathcal{C}_{\mathrm{graph}} \longleftarrow \mathrm{TopKChunks}(g_y, \mathbf{v}_q, k_{\mathrm{graph}})$\;
            \For{$c \in \mathcal{C}_{\mathrm{graph}}$}{
                $\mathcal{E}_{\mathrm{graph}} \longleftarrow \mathcal{E}_{\mathrm{graph}} \cup \mathrm{Entities}(c)$\;
                $\mathcal{R}_{\mathrm{graph}} \longleftarrow \mathcal{R}_{\mathrm{graph}} \cup \mathrm{Relations}(\mathcal{E}_{\mathrm{graph}}, g_y)$\;
            }
        }
    }

    $\mathcal{C}_{\mathrm{final}} \longleftarrow \mathcal{C}_{\mathrm{vec}} \cup \mathcal{C}_{\mathrm{graph}}$\;
    $context \longleftarrow \mathrm{concat}(\mathcal{C}_{\mathrm{final}}, \mathcal{E}_{\mathrm{graph}}, \mathcal{R}_{\mathrm{graph}})$\;

    $rep \longleftarrow \mathsf{LLM}.\mathrm{generate}(q, context)$\;
    \Return{$rep$}\;
}
\end{algorithm}

In terms of complexity, query embedding and classification are linear in the query length, vector and graph retrieval require $O(N_v d + n_y d)$ with brute-force search (or $O(\log N_v + \log n_y)$ with approximate nearest neighbor search), and entity expansion costs $O(k_{\mathrm{graph}}(\bar e + \bar r))$, where $\bar e$ and $\bar r$ denote the average numbers of entities and relations per chunk. The LLM generation cost $C_{\text{gen}}(L)$ usually dominates in practice. A key consideration is that large vector stores or dense subgraphs can significantly increase retrieval overhead, while excessive entity expansion may inflate the context length and thus raise the LLM cost. Hence, careful tuning of $k_{\mathrm{vec}}$, $k_{\mathrm{graph}}$, and context size is crucial to balance efficiency and answer quality.

\section{Dataset and Experimental Result}
\subsection{Dataset and Implementation Environment}

We compiled two distinct datasets to support our study. The first dataset comprises \textbf{\textit{1,319}} question–answer (Q\&A) pairs, carefully curated to evaluate the chatbot’s ability to address inquiries related to academic regulations. The second dataset contains \textit{3,256} questions developed for training a FastText-based classification model. This classification dataset was divided into training and testing subsets using an \textit{80:20} split. Most of the data were sourced from the official departmental websites of Can Tho University\footnote{\url{https://www.ctu.edu.vn/don-vi-truc-thuoc.html}} and other reputable sources. To ensure reliability, all datasets underwent rigorous manual verification. Furthermore, all responses were authored by academic support experts at Can Tho University. The training data and source code of our implementation are publicly available on GitHub\footnote{\url{https://github.com/tamB2203579/CatRAG}}. The experiments were conducted on a computer equipped with an Intel(R) Core (TM) i5-12500H 4.5 GHz processor, 16 GB of RAM, 6 GB of vRAM and running Windows 11 OS. Next, we will present the experimental results of our approach in the subsequent subsection.

\begin{table}[H]
\centering
\scriptsize
\caption{ Collected Dataset for REBot.}
\begin{tabular}{|c|l|c|c|}
\toprule
        \tiny{ID}
        & Name
        &  Number of
        & Note
        
        \\
        & 
        & Samples
        & 
        
        \\
 \midrule   
         \multicolumn{4}{|c|}{\textbf{Specialized Domains Classification}}\\
  \midrule
        1 &  Study and Training & 474 & Học tập và rèn luyện
        \\
        2 &  Education & 611 & Đào tạo
        \\
        3 &  Domitory & 655 & Ký túc xá 
        \\
        4 &  Discipline and Scholarships & 477 & Khen thưởng và kỷ luật
        \\
        5 &  Graduation & 485 & Tốt nghiệp
        \\
        6 &  Other & 550 & Khác
        \\
\midrule  
         \multicolumn{2}{|c|}{Total:}     &  3,252 &\\

\midrule   
         \multicolumn{4}{|c|}{\textbf{REBot Q\&A Evaluation}}\\
  \midrule 
    
        1 &  Truth & 909 & {\tiny Knowledge \textbf{related} to CTU academic regulation}\\ 
        2 &  Other & 404 & {\tiny Knowledge \textbf{outside} of CTU academic regulation}    \\
 \midrule   
               \multicolumn{2}{|c|}{Total:}     &  1,313 &\\
\bottomrule
\end{tabular}
\label{lb-Dataset}
\end{table}


\subsection{Quantitative Results}
We instruct REBot motivated from two architectural variants: the standard RAG and the graph-enhanced CatRAG (see Figure \ref{lb-combined-threshold}). As previously outlined, (1) the system utilizes phobert-base-v2 as the embedding model and retrieves the top five most relevant chunks to provide contextual input for the generative model. (2) For the CatRAG variant, a similarity threshold of 0.7 is applied when leveraging the knowledge graph to ensure semantically relevant node connections. The experiments are conducted using two generative AI models: \textit{gpt-4o-mini and mistral-small-2506}. To assess the performance of both architectures under different threshold configurations, we employ four core evaluation metrics: Accuracy, Precision, Recall, and F1 Score. Among these, the F1 Score is prioritized as the primary criterion for model selection due to its balanced integration of Precision and Recall.

\begin{table*}[t]
\centering
\scriptsize
\caption{Comparison of CatRAG (CR) and RAG (R) across thresholds}
\BlankLine
\begin{tabular}{|c|c|c|c|c|c|c|}
\toprule
Provider & Threshold & Accuracy & Precision & Recall & F1 Score & Eval. Time \\
\midrule
 &  & (CR \,|\, R) & (CR \,|\, R) & (CR \,|\, R) & (CR \,|\, R) & (CR \,|\, R) \\
\midrule
\multirow{3}{*}{OpenAI} 
& 0.6 & 98.48 | 98.17 & 99.33 | 98.78 & 98.46 | 98.56 & 98.89 | 98.67 & -- | -- \\
& 0.7 & 95.58 | 93.75 & 95.11 | 92.32 & 98.39 | 98.46 & 96.72 | 95.29 & 2h10m12s | 2h10m12s \\
& 0.8 & 88.58 | 86.44 & 84.87 | 81.65 & 98.20 | 98.26 & 91.05 | 89.19 & -- | -- \\
\midrule
\multirow{3}{*}{Mistral AI} 
& 0.6 & 98.25 | 98.55 & 99.11 | 99.33 & 98.35 | 98.57 & 98.73 | 98.95 & -- | -- \\
& 0.7 & 95.66 | 93.99 & 95.33 | 92.67 & 98.28 | 98.47 & 96.79 | 95.48 & 1h52m47s | 1h47m17s \\
& 0.8 & 85.00 | 82.86 & 79.76 | 76.42 & 97.95 | 98.14 & 87.92 | 85.93 & -- | -- \\
\bottomrule
\end{tabular}
\label{lb-combined-threshold}
\caption{Computation Time of CatRAG and RAG (average of 10 runs).}
\begin{tabular}{|c|c|c|c|}
\toprule
\scriptsize{ID} & Model Name & CatRAG (s) & RAG (s) \\
\midrule
1 & gpt-4o-mini (OpenAI)          & 7.2577 & 5.1076 \\
2 & mistral-small-2506 (Mistral)   & 7.0232 & 4.7692 \\
\bottomrule
\end{tabular}
\label{lb-ComputationTime}
\end{table*}

The results of both architectures are presented in Table \ref{lb-combined-threshold}. Among the configurations, CatRAG combined with gpt-4o-mini demonstrates the most impressive performance, consistently outperforming its RAG counterpart across all threshold settings. Notably, it achieves a near 98.9\% F1 Score at a threshold of 0.6, while still maintaining over 90\% F1 Score at a threshold of 0.8, indicating strong robustness across different sensitivity levels. In contrast, Mistral AI exhibits more variability across thresholds. While the performance at thresholds 0.6 and 0.7 is comparatively lower, it achieves a notable improvement at threshold 0.8, suggesting that the model performs better when operating under stricter filtering conditions. Note that, we conduct a gridsearch from $0.1$ to $1.0$. Then, $[0.6,0.8]$ obtained the high accuracy.

\begin{table*}[t]
\centering
\footnotesize
\scriptsize
\caption{Results of Specialized Domain/Topic Classification.}
 \begin{tabular}{|c|c|c|c|c|c|c|c|c|}
\toprule
        \scriptsize{ID}
        & Models
        &  Parameters
        &  Acc
        &  Prec
        &  Rec
        &  F1
        & \multicolumn{2}{c|}{Time (seconds)}
        \\
       
        & 
        &  
        &  (\%)
        &  (\%)
        & (\%) 
        & (\%)
        & Training
        &Testing
        \\
\midrule
    1 & Multinomial NB & $alpha=0.1$ & 87.12  &  87.16 & 87.12 & 86.78 & 0.03 & 0.01
        \\
        2 &  KNN & $k=101$ & 83.74  &  85.61 & 83.74 & 82.07 & 0.03 & 0.183
        \\
        3 &  Logistic Regression & {\scriptsize $C=10$,$tol=0.0001$ } & 93.87  &  93.95 & 93.87 & 93.81 & 1.16 & 0.009
        \\
        4 &  SVM & {\scriptsize $C=1$,$tol=0.0001$, $loss=hinge$ } & 91.41  &  91.23 & 91.41 & 91.02 & 0.16 & 0.010
        \\

         5 &  Random Forest &  {\scriptsize $n\_estimators=300$} & 94.79 &  94.92 & 94.79 & 94.66 & 0.79 & 0.186
        \\
         6 &  FastText & {\tiny $epoches=100, Ngrams=3$}  & 95.75  &  95.76 & 95.75 & 95.67 & 2.2775 & 0.056\\
\bottomrule
\end{tabular}
\label{lb-ResultOfClassification}
\end{table*}

Table \ref{lb-ComputationTime} shows that CatRAG consistently incurs longer response times than standard RAG due to graph-based retrieval and node similarity computation. For instance, with gpt-4o-mini, CatRAG averages 7.26 s per query versus 5.11 s for RAG, and with mistral-small-2506, 7.02 s versus 4.77 s. Despite this $\sim$2 s overhead, latency remains acceptable for real-time use given CatRAG’s enhanced retrieval capability.  
Table \ref{lb-ResultOfClassification} reports classification results, where Random Forest (94.79\% accuracy, 94.66\% F1) and Logistic Regression (93.87\%, 93.81\%) perform strongly among traditional models, but KNN lags (83.74\%, 82.07\%). FastText achieves the best performance (95.75\% accuracy, 95.67\% F1), demonstrating superior ability to capture semantic features while maintaining efficiency.

\subsection{Qualitative Result}
The questions and answers are expressed with clarity and effectively address users’ needs. Illustrative examples are presented as follows:

{\tiny
\begin{itemize}
    \item Question 1:
    \textbf{Input:} ~~How many semesters are there in an academic year?\\
    \textbf{Output:} \textit{An academic year at Can Tho University consists of three semesters, called semester 1, semester 2, and semester 3. Each semester lasts 15 weeks, including time for teaching, examinations, and result processing.}
 
    \item Question 2:\\
    \textbf{Input:} ~~What are the foreign language graduation requirements for students?\\
    \textbf{Output:} \textit{The foreign language graduation requirements at Can Tho University include:}
    \begin{enumerate}
        \item \textit{\textbf{Meeting language proficiency standards}: Students must reach the required foreign language level according to their training program. Specifically:}
        \begin{itemize}
            \item \textit{Students of cohort 45 and later must have a foreign language certificate at level 3/6 according to the Vietnamese 6-level language proficiency framework or equivalent.}
            \item \textit{Students of cohort 44 and earlier must have a French certificate level 2 or above, or an equivalent English certificate.}
        \end{itemize}
        \item \textit{\textbf{Participation in intensive language programs}: Students need to join intensive language programs organized by the university to improve their foreign language skills to the minimum required level.}
        \item \textit{\textbf{Exemption from language courses}: Some students may be exempt from basic language courses if they hold an appropriate certificate or passed the language proficiency test at the beginning of the course, as decided by the language proficiency council.}
    \end{enumerate}
    \textit{In summary, to graduate, students must complete the foreign language requirements of their training program.}
\end{itemize}
}

In this paper, we distilled four key questions reflecting practical concerns of Can Tho University students. The responses were largely accurate and contextually enriched, though certain details require further refinement. Overall, the information is sufficient for users to draw informed conclusions. Finally, we present the REBot interface and discuss its broader implications.


\subsection{Application and Discussion}

\begin{figure}[ht]
 \begin{center}
      \includegraphics[scale=0.1]{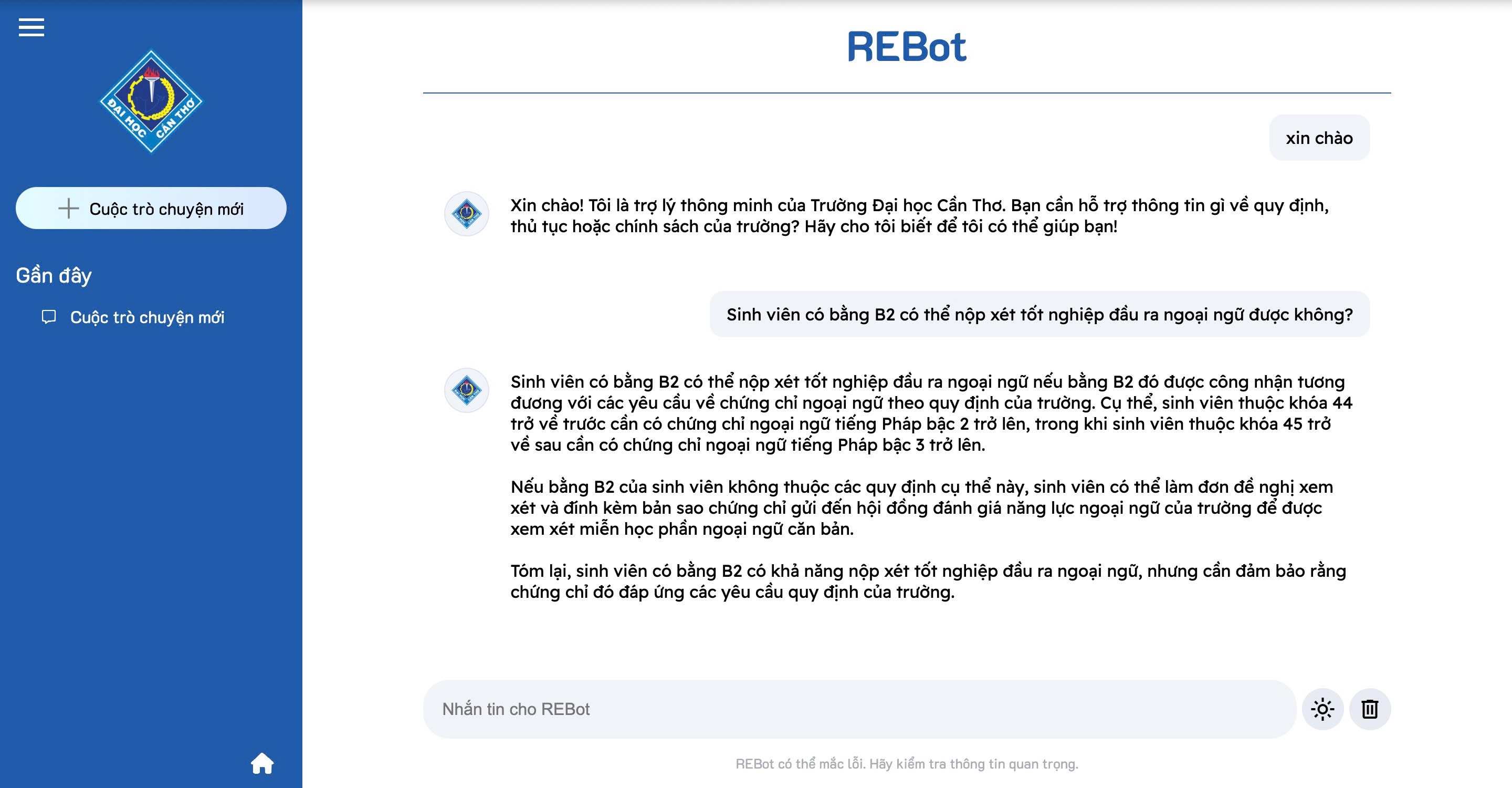}
 \end{center}
 \caption{The Interface of our REBot System}
 \label{REBotInterface}
 \end{figure}

After training the AI models, we deploy the framework on a web-based platform integrated with the REBot interface (Figure~\ref{REBotInterface}). The system is implemented using modern libraries and frameworks, including React (19.0.0), ViteJS (6.3.1), and Bootstrap (5.3).

REBot demonstrates effective real-time conversational interaction with students; however, several limitations remain. (1) GraphRAG-based retrieval can introduce excessive context, leading to incoherent or inaccurate responses. (2) dataset classification requires improvement for handling long-context queries. (3) the current average response time of about seven seconds is a significant bottleneck. As future work, we plan to optimize retrieval strategies and reduce latency, while expanding the dataset vertically to capture finer-grained regulatory details within each academic domain (e.g., specific course requirements, exceptions, procedural workflows). Finally, we will be refining the domain-specific dictionary, enhancing classification models, and incorporating adaptive retrieval with reranking for contextual precision.

\section{Conclusion}
This study presents the effectiveness of a retrieval-augmented chatbot system using standard Retrieval-Augmented Generation (RAG) and knowledge-graph-enhanced GraphRAG to provide accurate and detailed responses to student questions at Can Tho University. Quantitative tests indicate that CatRAG, combined with a compact generative model, achieves top performance with an F1-score 98.89\%, while keeping computation times practical for real-time use. Qualitative reviews confirm the system's value in offering clear, in-depth explanations that go beyond simple queries to improve user understanding. By incorporating advanced embedding techniques and knowledge graphs, this approach beats traditional methods and lays the groundwork for flexible, domain-focused AI helpers in education.
\\

\noindent\textbf{Acknowledgements:}
This research is part of Can Tho University’s scientific research program under the code THS2025-69.

\bibliographystyle{splncs04}
\bibliography{mybibliography}


\end{document}